\documentclass[conference]{IEEEtran}
\IEEEoverridecommandlockouts
\usepackage{cite}
\usepackage{float}
\usepackage{adjustbox}
\usepackage{times}
\usepackage{enumitem}
\usepackage{graphicx}
\usepackage{subfigure}
\usepackage{graphicx}
\usepackage{draftwatermark}
\SetWatermarkText{Accepted in SoSE 2019}
\SetWatermarkScale{0.2}
\begin{document}

\title{HRC-SoS: Human Robot Collaboration Experimentation Platform as System of Systems}
\author{{Celal Savur$^{1}$ \qquad Shitij Kumar$^{2}$ \qquad Sarthak Arora$^{3}$ \qquad  Tuly Hazbar$^{4}$ \qquad Ferat Sahin$^{5}$}
\\
\IEEEauthorblockA{Department of Electrical and Microelectronic Engineering \\
Rochester Institute of Technology\\
Rochester, NY, 14623, USA \\
\{cs1323$^{1}$,spk4422$^{2}$,sa9472$^{3}$,tmh6831$^{4}$,feseee$^{5}$\}@rit.edu }
}
\maketitle
\begin{abstract}
This paper presents an experimentation platform for human robot collaboration as a system of systems as well as proposes a conceptual framework describing the aspects of Human Robot Collaboration. These aspects are \emph{Awareness}, \emph{Intelligence} and \emph{Compliance} of the system. Based on this framework case studies describing experiment setups performed using this platform are discussed. Each experiment highlights the use of the subsystems such as the digital twin, motion capture system, human-physiological monitoring system, data collection system and robot control and interface systems. A highlight of this paper showcases a subsystem with the ability to monitor human physiological feedback during a human robot collaboration task.      

\end{abstract}


\section{Introduction}
\label{sec:introduction}
As the production requirements are constantly changing, Human-Robot Collaboration (HRC) is receiving a high interest in industry. While the benefits of industrial robots working in isolation are still valued when the productivity is the main goal in production. The shortcomings of these pure robotic cells become more apparent when flexibility is also valued along with short production cycles and customized product demands. This shift in production interest suggests Human Robot collaboration as a viable alternative. 
 \footnotetext{\footnotesize{Accepted in \textit{14th International Conference on System of Systems Engineering (SoSE)}, May 19-22 , 2019 Anchorage, Alaska, USA}}
The concept of HRC is not new, there are many examples of HRC applications that are revolutionizing a diversity of fields and one in particular is manufacturing.
An aspect of HRC is bringing forth the ability for robots to perform new tasks from natural human instructions, learn new tasks from a human expert demonstration and work with humans in the same shared workspace. This applies especially to humans who are domain experts but not robotics experts \cite{Ko2015} and are working alongside robots without hindering each others productivity \cite{kumarDynamicAwarenessIndustrial2018}. An example is a robot performing tasks collaboratively with a human teammate, where both the skills of the human and the robot can complement each other to accomplish a task that neither can achieve alone \cite{Thomas2016}. While the number of ways a robot can collaborate with a human are limitless, this type of collaboration introduces new challenges to robotics research for industrial settings and demands for a well defined industrial standards. 

The major challenges of Human Robot Collaboration (HRC) in industry are human safety, human trust in automation, and productivity \cite{kumarDynamicAwarenessIndustrial2018}. Human safety has always been the primary concern in robotics. One main aspect is injuries due to human-robot collision. Different strategies have been introduced to ensure human safety. One is implementing  physical and electronic safeguards according to industrial standards \cite{ISO2011}\cite{ISOTS15066}. The lack in new strategies and approaches within human robot collaboration where there are less standards available to implement complex protection schemes create a demand for a new category of robots called \emph{collaborative robots} or \emph{cobots}. These robots are purposely designed to work in direct cooperation with humans in a defined workspace by lowering the severity and risks of injury due to collision. 

\begin{figure}[t!]
	\centering
	\includegraphics[width=6.5cm,keepaspectratio]{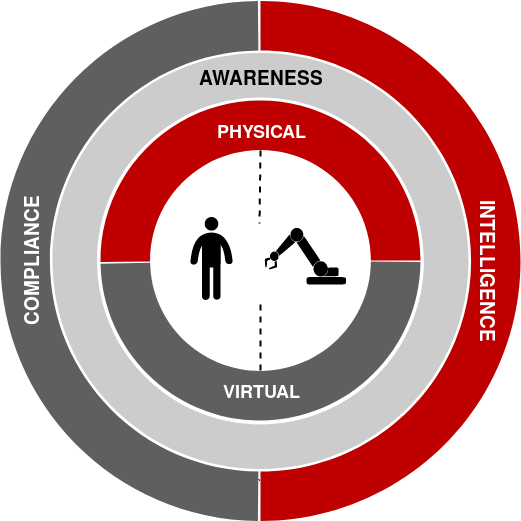}
	\caption{A system diagram for the proposed conceptual framework for a Human Robot Collaboration as System of System. It highlights the 3 aspects of an HRC system : \emph{Awareness}, \emph{Intelligence} and \emph{Compliance}.}
	\label{fig:hrc_conceptual_framework}
\end{figure}

Human trust in automation is about managing human expectations and how comfortable the human is in sharing the robot workspace. Even though \emph{cobots} decrease the risk of injury, any form of physical collision decreases the human trust in automation. Thus, collision avoidance strategies such as stopping or reducing speeds while the human is in the operating workspace of the robot, are implemented to avoid any kind of human-robot collision or degrade the trust relationship between them. 

Safety and Trust both positively affect the productivity of the robot, however efficient robot motion and anticipation of human actions and presence can also ensure faster execution of tasks. In industry cycle times are important. There is no doubt a fully automated system would provide the highest productivity. However as soon as the production requirements change, flexibility of production and swiftness needs to be optimized. 

Thus research in understanding and analysing HRC setups that ensures human safety, builds human trust in automation while optimizing the productivity are required. With that objective in mind, a Human Robot Collaboration framework as a System of Systems HRC-SoS is proposed. In our opinion, for any HRC system three main aspects are to be considered: \emph{Awareness}, \emph{Intelligence} and \emph{Compliance} which were introduced briefly in our previous work \cite{kumarDynamicAwarenessIndustrial2018}. These aspects are discussed in detail in the further Sections and case studies validating this framework have been presented as results. A conceptual system diagram for the proposed HRC-SoS is shown in Figure \ref{fig:hrc_conceptual_framework}.

\section{Proposed HRC-SoS Framework}
\label{sec:approach}
The three aspects of a Human Robot Collaboration setup from a System-of-Systems perspective can be identified as three subsystems : \emph{Awareness}, \emph{Intelligence} and \emph{Compliance}.  

\emph{Awareness} of a robot is the level of robot perception using sensors. \emph{Intelligence} is the robot action and behavior. \emph{Compliance} is about the robot managing human expectation, human control over the robot and efficient communication and feedback between human and robot. A few examples of each are listed as follows:
\begin{list}{\textbullet}{\leftmargin=0.5em}
	\item  \textbf{Awareness} addresses the perception of environment and human safety issues by using sensors such as air-curtains, 2D scanning lidars, cameras, pressure pads, buttons etc. Collaborative Robots are an example of more aware industrial robot with inbuilt force/torque sensing actuators. 
    \item \textbf{Intelligence} - \textit{Computer vision }: object recognition, tracking, visual servoing, \textit{Motion Planning}: navigation, trajectory planning and generation, obstacle avoidance, \textit{Task Planning and Execution}: task scheduling, task management and anticipating human actions.	\textit{Learning}: Reinforcement Learning, learning from demonstrations or other forms of learning for modeling Robot Behavior.
    \item  \textbf{Compliance}- \textit{Human Control}: speech based, gesture based, teaching by demonstration and ability to free-drive. \textit{Communication}:  Efficient and ergonomic human-machine interfaces, robot current and future actions conveyed to the human ; and human feedback to the robot verbal, haptic or physiological.  
\end{list}

\subsection{\textbf{Awareness in HRC-SoS}}
\label{sec:Awareness}
 With the onset of human robot collaboration (HRC), the interaction between the operator and the robot have become extremely human-centric. For any interaction to safely occur, information associated with human/operator position with-respect-to the robot must be present. Usually, these scenarios are rife in factory floors and indoor environments. Therefore, the use of exteroceptive sensor systems such as \cite{optitrack} have enabled complete human tracking including bio-mechanical information. 
 
 Sensors results in perception, and perception defines the \emph{Awareness} of a robot.  Sensor limitation dictates the level of \emph{Awareness}. It must be noted that these sensing systems are setup and mounted in the robot's environment and usually require calibration routines and planning of sensor placement around the concerned volume of operation. However, due to the densely occluded nature of indoor environments and factory floors, occlusion becomes inevitable. To alleviate this problem the use of exteroceptive sensors affixed to the robot is a viable option.

As it can be verbose and confusing to refer to the aforementioned sensors with their designated terms. For convenience, the systems can be divided into two categories similar to virtual-reality (VR) tracking systems. When the tracking system is completely self-contained within the VR headset it is referred to as ``inside-out" tracking. When the tracking system is completely external to the VR headset it is referred to as ``outside-in" tracking. Similarly, when a sensing system is affixed on the robot it would be convenient to express the system as an ``inside-out" sensing system from the robot's perspective and ``outside-in" from the sensors mounted in the environment. A similar idea was proposed in \cite{kumarDynamicAwarenessIndustrial2018} where the author(s) classified ``inside-out" \& ``outside-in" as intrinsic \& extrinsic sensing systems respectively (see Figure \ref{fig:sensing}). 

Currently in research and industry the ways of perception from robots perspective can be categorized as follows:
\begin{figure}
	\centering
	\begin{subfigure}[]
	{
			\centering
			\includegraphics[width=0.45\linewidth,keepaspectratio]{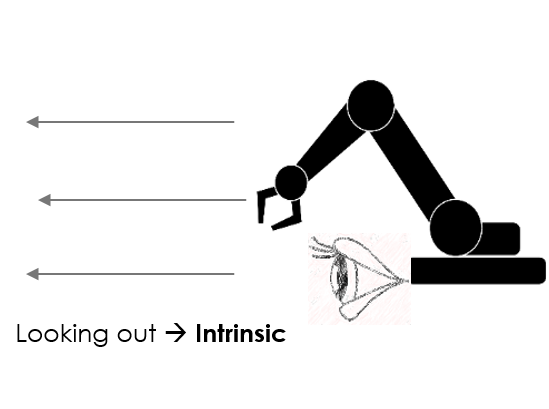}
 			\label{fig:intrinsicSensing}
 			}
	\end{subfigure} 
	\begin{subfigure}[]{
			\centering
			\includegraphics[width=0.45\linewidth,keepaspectratio]{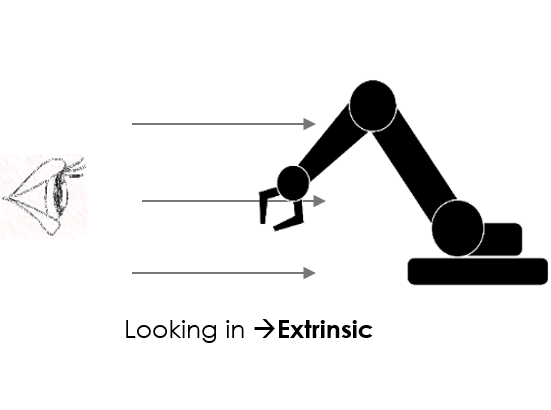}
 			\label{fig:extrinsicSensing}
 			}
	\end{subfigure}
	\caption{(a) A representation of intrinsic sensing i.e. looking out  (b) A representation of Extrinsic sensing i.e. looking in }
	\label{fig:sensing}
\end{figure}

\subsubsection{\textbf{Intrinsic Sensing}}
Theses can be further categorized as follows:
\begin{list}{\textbullet}{\leftmargin=0.5em}
		\item \textbf{Distance based} - e.g. Air curtains, 2D lidars \cite{marvelImplementingSpeedSeparation2017}, pressure pads are example of statically aware perception. Other sensors on robots such  as Time-of-Flight (ToF) single unit lidars, end-of-arm-tool (EOAT) cameras etc. are dynamic.
		\item \textbf{Tactile Based} - Using a pressurized Skin \cite{Cirillo2016}, capacitive touch \cite{schleglVirtualWhiskersHighly2013} and buttons on robot links or EOAT.
		\item \textbf{Inbuilt Force/Torque Sensors in the Actuators of the Robot} - this is what has led to the development of collaborative robots and physical HRI \cite{Geravand2013}. 
	\end{list}  
\subsubsection{\textbf{Extrinsic Sensing}}
Examples of Extrinsic sensors are Cameras (inaccurate and inexpensive), Kinects/3D cameras (somewhat accurate and inexpensive) used in \cite{flaccoDepthSpaceApproach2015}, 3D Lidars (accurate and very expensive), Motion Capture Systems (very accurate and very expensive) \cite{safeeaMinimumDistanceCalculation2019}\cite{CASE2019_paper}. 




In the case studies and experiments conducted, the \textit{Awareness} subsystem has both extrinsic ans intrinsic sensors for monitoring the environment of human-robot shared workspace. The sensors used are : 
\begin{list}{\textbullet}{\leftmargin=0.5em}
    \item OptiTrack 3D tracking system, The OptiTrack is a motion capture system which tracks the moving objects by using Flex 13 cameras which can reach millimetre precision in real-time \cite{optitrack}. Human movement tracking is crucial for a human-in-the-loop system. The tracking information is useful in generating a virtual world representation (`Digital Twin') of the moving agents which are the human, robot and objects in the physical world.
    \item Multiple RGB-D cameras such as Microsoft Kinects and Intel Real Sense are used to generate a 3D pointcloud of the workspace, similar to the approach used in \cite{flaccoDepthSpaceApproach2015}.
    \item Time-of-Flight Sensor Arrays are used for speed and separation monitoring (SSM) \cite{marvelImplementingSpeedSeparation2017} and developing a human-position estimation system using only `intrinsic' sensing \cite{CASE2019_paper}.
\end{list}

\subsubsection{\textbf{Digital Twin}}
`Digital Twin' can be defined as a bridge between real/physical  and virtual-digital world. In the context of human robot collaboration, the human, the robot and the environment have  digital counterparts that accurately replicate their states in the real/physical world. As represented in Figure \ref{fig:hrc_conceptual_framework} the Awareness of our conceptual system is comprised of the information presented in the physical and virtual world. A key component of the implementation of the digital twin is the use of `physics engines' that allows accurate virtual world representation of the physical world. Although the virtual world does not account for the uncertainties of the real world, the observable nature of the virtual representation is proven to be powerful testing approach for algorithms and HRC experiments. Other possible applications of the digital twin are process evaluation before, during or after the execution and real-time scene assessment for revealing hidden aspects of the collaboration via visualization or other human intelligible feedback \cite{Cichon2017} \cite{safeeaMinimumDistanceCalculation2019}. 

The `physics engines' integrated in the HRC-SoS are listed as follows : 
\begin{list}{\textbullet}{\leftmargin=0.5em}
	\item \textbf{PyBullet Interface}- A physics engine simulation for performing and representing the robot dynamics and kinematics \cite{pybullet2018}. It has functionalities to integrate with various VR applications and also has deep learning and reinforcement learning platform compatibility. Thus allowing future research for mixed-reality and virtual reality interfacing during an HRC experiment. This subsystem interface can be used to virtually represent the physical world environment and use its functionalities to calculate robot dynamics, and kinematics. In a case study the PyBullet engine, was used to create a self-occlusion detection for the sensors mounted on the robot \cite{CASE2019_paper}.
	\item \textbf{V-REP Interface} Virtual Robotics Experimentation Platform \cite{rohmerVREPVersatileScalable2013} is a user friendly 3D simulation platform to create workspace and interactions of Robot. This platform can be used to create virtual interactions with the simulated robot, that can affect the real robot behavior \cite{kumarDynamicAwarenessIndustrial2018} \cite{CASE2019_paper}.
	\item \textbf{ROS Interface(RVIZ/Gazebo/MoveIT)} - Robot Operating System (ROS) \cite{Fat1946} is crucial for fast prototyping and using the vast knowledge base to implement algorithms. ROS can be used to generate and receive data from the generated \textit{Awareness} data of the system. As this data can be available to ROS, 3D environments such as Gazebo, RVIZ and MoveIt can be used to represent and process data.
\end{list}    

\subsubsection{Communication Layer of HRC-SoS}
\label{sec:communication}
In order for the information of available from the sensing systems to be efficiently and timely communicated and also stored, ZeroMQ messaging protocol along with Robot Operating System is used. 

\subsection{\textbf{Intelligence in HRC-SoS}}
\label{sec:intelligence}
 \textit{Intelligence} subsystem is responsible for actionable commands mainly to the robot and extracting information from sensor data. As it can be seen in Figure \ref{fig:hrc_conceptual_framework} it communicates with \textit{Awareness} and \textit{Compliance } subsystems. In this section three different case studies that have been implemented are discussed and the intelligence aspect highlighted.
 
 \subsubsection{\textbf{Case 1: Human Pose Estimation via Intelligent Awareness}}
 \label{sec:case1}
 \begin{figure}
 \centering
  \includegraphics[width=6.5cm,keepaspectratio]{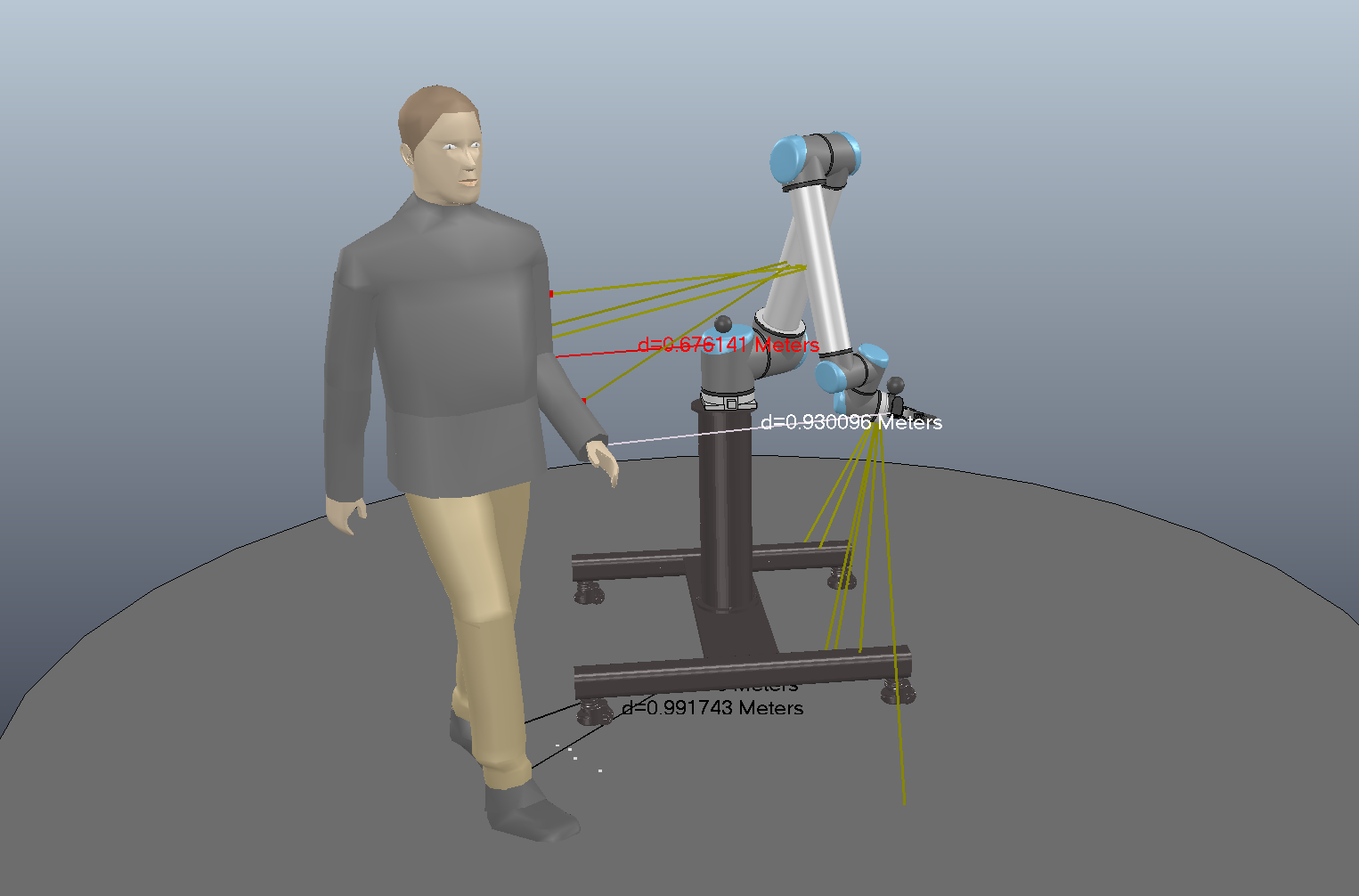}
  \caption{UR10 robot with time--of--flight sensors mounted on each link, rays are shown when hit. The observation from each lidar are used to estimate and track the human operator in the robot workspace.}
  \label{fig:cover}
\end{figure} 
 For any compliance strategy to be successful, a robust awareness strategy is extremely important. The awareness scheme must provide an optimum amount of information that can conveniently be interpreted into actionable information by a compliance algorithm. Therefore, eliciting information associated with the human's location (in the context of robot workspace), from spatio-temporal sequences produced by the distance sensing devices mentioned above, is vital. Transmitting the raw data into the former has hitherto been a daunting challenge. Reaping the benefits provided by modern machine learning techniques such as deep learning and advanced motion tracking systems such as OptiTrack; a mapping between raw sensing data and human position can potentially facilitate the compliance policy. The setup for this case study is shown in Figure \ref{fig:cover}.
 
 \subsubsection{\textbf{Case 2: Speed and Separation Monitoring using ToF Laser Ranging Sensor Arrays}}
 \label{sec:case_2}
  \begin{figure*}[t!]
    \centering
     \includegraphics[width=0.95\textwidth]{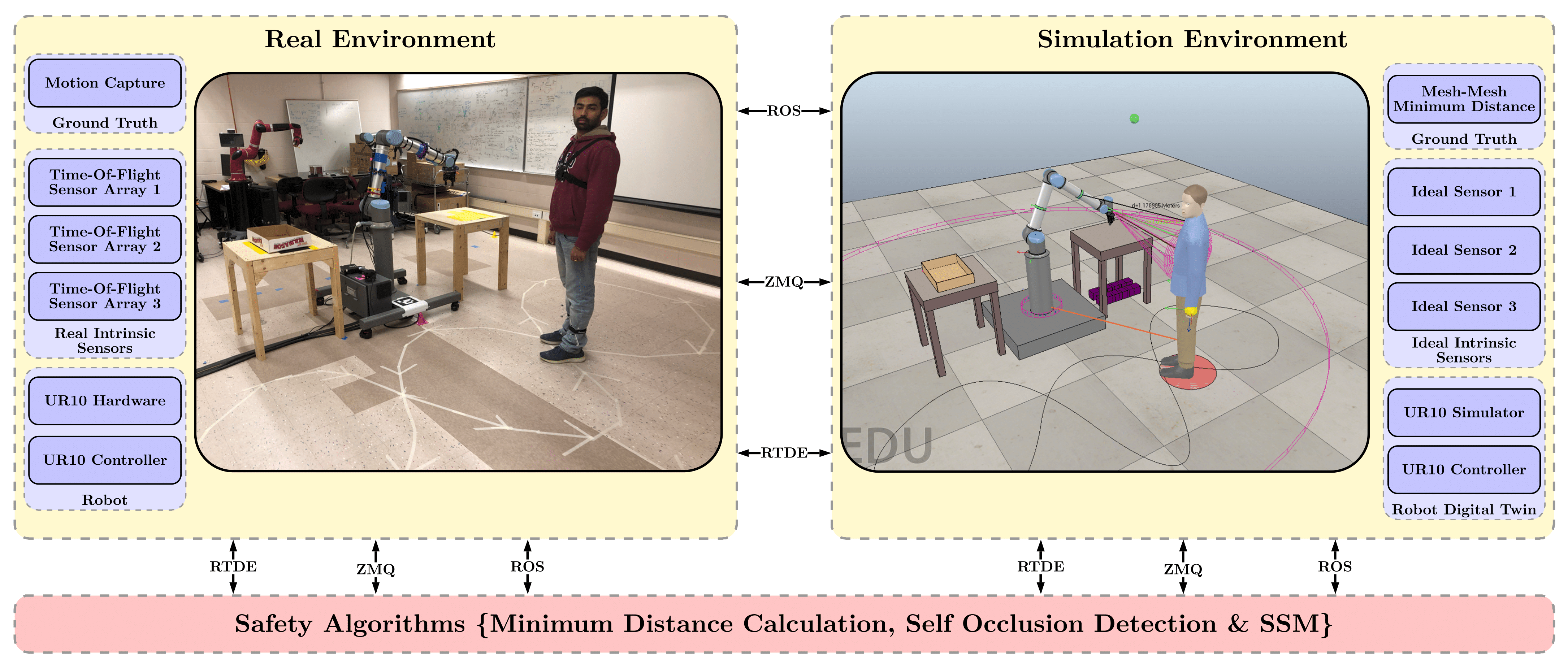}
    \caption{A schematic of the system used to implement, validate and test the proposed SSM safety configuration(s). The transport layer for communication between different subsystems such as the real and virtual environments is built using ZeroMQ, RTDE and ROS \cite{CASE2019_paper}.}
    \label{fig:new-setup-diag}
\end{figure*}
This case study refers to the research presented in \cite{kumarDynamicAwarenessIndustrial2018} and \cite{CASE2019_paper}. In this research there were experiments performed for Safety, Performance and Productivity of variations of Speed and Safety Monitoring setups for minimum distance(s) calculated using the \textit{Physical} ToF rings mounted on the robot, the \textit{Virtual} minimum distance from center of the links to the human/obstacle measured using motion capture and V-REP environment,  and 2D scanning \textit{Lidar}. In \cite{CASE2019_paper}, several arguments were presented to demonstrate the similarity between the simulated and the physical versions of the ToF sensor arrays. An ISO compliant \cite{ISOTS15066} safety algorithm was also presented and implemented in \cite{CASE2019_paper}. A schematic of the experiment and validation setup is shown in Figure \ref{fig:new-setup-diag}.

\subsubsection{\textbf{Case 3:  Plan Execution of a Shared Task in a Human-Robot Team}}
\begin{figure}[ht]
	\centering
	\includegraphics[width=8.5cm,keepaspectratio]{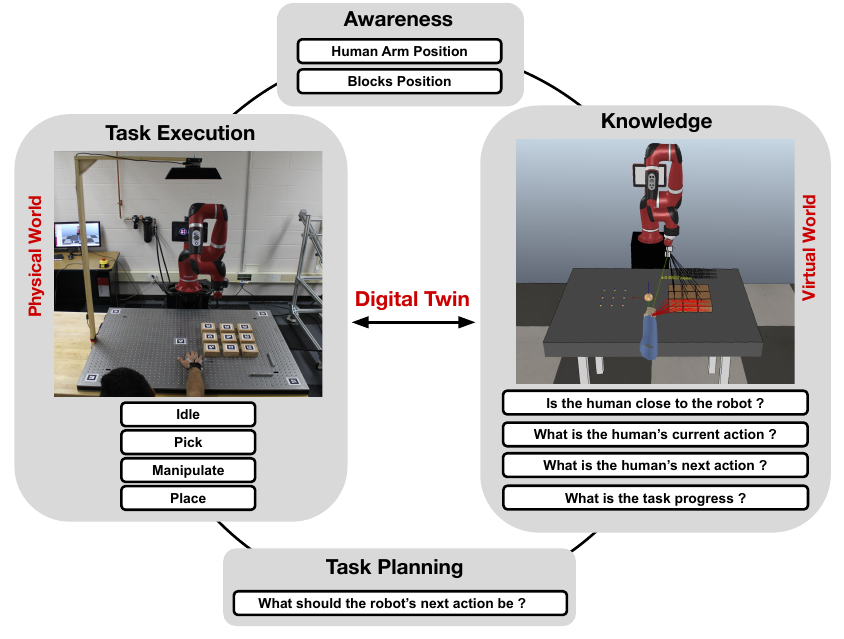}
	\caption{Overview of the Collaborative Plan Execution Framework}
	\label{fig:Plan_Framwork}
\end{figure}

One aim of the HRC research is to enable the robot to not only share the workspace with human, but also preform a joint task with a human counterpart. When a team of two humans work together, they display a high level of action coordination. They alter their plans and select their actions appropriately. In order to consider the robot as a team-mate rather than a tool, it should exhibit this level of intelligence in a team. In this case study, the human-robot team is given a set of toy building blocks with a goal to achieve a final arrangement fluently. An overview of our approach is shown in Figure \ref{fig:Plan_Framwork} \cite{SMC2019_Tuly}. The following explains how we integrated both robot awareness and robot intelligence explained in Section  and illustrate how the digital twin concept helped in implementing such framework. 
\begin{list}{\textbullet}{\leftmargin=0.5em}
     
\item \textbf{System Perception and Robot Awareness:} For the robot to be a better collaborator, it should be aware of the human partner and his/her actions and the progress of the shared task. Using sensors the robot will have a perception of the position of the human arm and blocks at all times. These two piece of information are crucial to our HRC scenario for safety and task planning purposes. Through perception the knowledge will be defined in the next section. 

\item \textbf{Forming Knowledge through Digital Twin:} Every element in the physical world that affects the HRC scenario has a counterpart in the virtual world that mirrors its current state. The input to the virtual world is the data of the perception component, and the output of the virtual world is the integrated knowledge of the human arm, the robot and the blocks. Figure \ref{fig:Plan_Framwork} lists the knowledge we are interested in for our scenario. 

\item \textbf{Robot Intelligence:} Given the set of knowledge, the robot needs to select on the fly a corresponding action from a set of possible known actions. There are three main types of decisions a robot can make: immediate decisions concerning human safety, proactive and reactive decisions regarding collaboration fluency and the task finial goal. A formalism that combines Concurrency with Hierarchical Finite State Machine (HFSM) is used to implement this behavior to account for situations where interrupting the normal behavior is necessary to respond to something more important.
 \end{list}
\subsection{\textbf{Compliance in HRC-SoS}}
\label{sec:compliance_system}
In the \textit{Awareness} and \textit{Intelligence} aspects of HRC, human is just a dynamic object in the environment, however, \textit{Compliance} of robot system tries to interpret actionable and human mental and physical state information from human physiological signal as a form of feedback and control. This can be further used in quantifying trust in automation and  evaluating human robot interaction. There is large body of research in classification and interpretation of human physiological signals for both control modality and identification of human state. In order to have such a system data collection and monitoring  become crucial in an HRC setup. The following section highlights some of the commonly used physiological signals and how they can be synchronized and recorded. 
\subsubsection{\textbf{Human, Physiological Signals}}
\begin{list}{\textbullet}{\leftmargin=0.5em}
	\item \textbf{Electroencephalogram (EEG)}  EEG has been used for error related potential, it has also been used to detect alpha activity, which determines attentiveness, stress, and other emotion. It can be argued that wearing an EEG cap while working can be uncomfortable. However, it must be noted that in industry workers can wear helmets or hats. With the advent of advance IoT systems wireless communication and small size factor of EEG equipment make it plausible to get such data.
	\item \textbf{Electrocardiogram (ECG)} measures the heart's electrical activity. ECG can be an indicator for physical and mental stress and fatigue. In case workers working in an industrial setup robot behavior can be adjusted based on the state of the operator. We hope this can increase trust in automation and also avoid any injuries from work exhaustion. Some of the medical devices that have been interfaced are Bitalino and BioRadio \cite{Ali2018}. 
	\item \textbf{Electromyogram (EMG)} as been used as an control input for basic robot interaction in many studies. A sense of control is very important for building the trust in  human robot interaction. Another example of EMG is using facial muscles to give information about sudden emotional changes or reactions.  Placement of these can be in safety glasses  worn by the operator \cite{Kulic2007},\cite{Gouizi2011},\cite{Savur2017}.
	\item \textbf{Electro-dermal Activity (EDA)} or Skin Conductivity (SC) or Galvanic Skin Response (GSR) is a measure of skin conductivity triggered by the human central nervous system. This signal has been used in emotion recognition, in lie detectors, and as an indicator of physical and mental stress \cite{Kulic2007},\cite{Rohrmann1999},\cite{Ali2018}.
	\item \textbf{Heart Rate (HR) and Heart Rate Variability (HRV)} is a signal that can be extracted from the ECG signal. This information can give the resting or active state of a human operator.
	\item \textbf{Pupil Dilation} is a measurement of pupil diameter changes. The pupil dilation can be caused by ambient or other light intensity change in environment. It also can be used to detect emotional change \cite{Bonifacci2015}.
\end{list}
\subsubsection*{\textbf{Lab Stream Layer (LSL)}}
For combining various device data in real time with that of the robot and environment state can be challenging. Hence we propose an interface subsystem in which we integrate data from all different devices using LSL layer. These are then interpreted into actionable input to the intelligence of the system. The Lab Stream Layer is a system for collection time series data over a local network with built-in time synchronization \cite{SCCN2018}. The LSL stream is nearly real-time and it is commonly used in biological signal collection system such as OpenBCI, Pupil Lab and g-tec systems. 

\begin{figure}[ht]
	\centering
	\includegraphics[width=8.5cm,keepaspectratio]{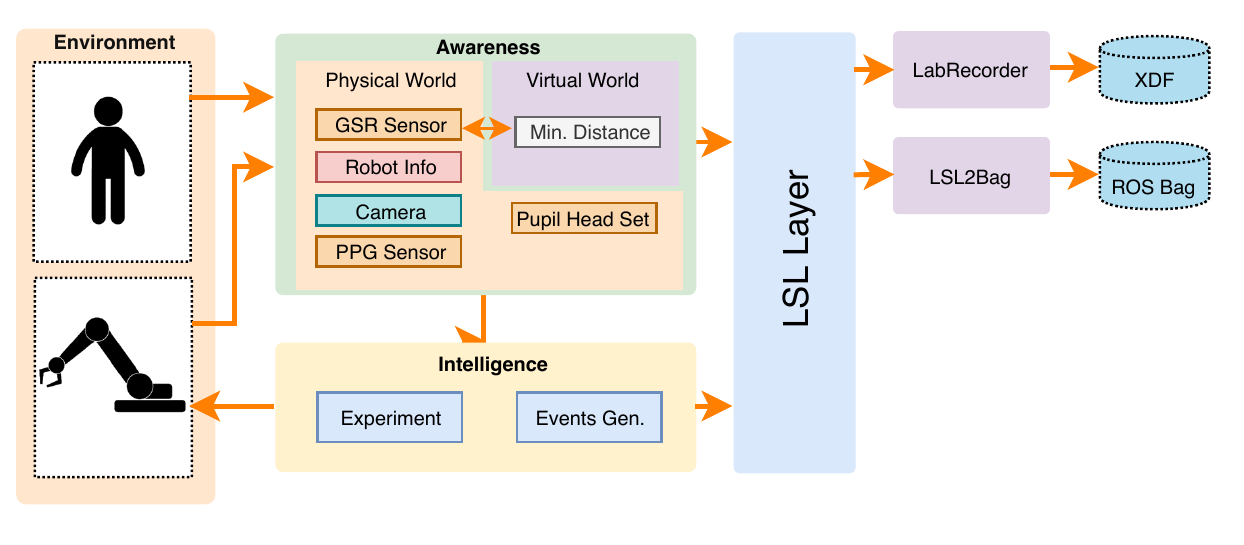}
	\caption{A schematic for monitoring and collection of human physiological signals during an HRC experiment.}
		\label{fig:Data-collection-system}
\end{figure}

The LSL system is central to the human physiological data acquisition system for this research. All the time series data will be transferred over LSL layer to be recorded in a database for storage. Each device will have its application/interface that retrieves signal in real-time and pushes the signal to an LSL stream. Since LSL ensures synchronous data collection and stream, it voids the need to perform time-synchronization. A schematic of the overall data collection system is shown in Figure \ref{fig:Data-collection-system}.

As can be seen, there are several devices in figure \ref{fig:Data-collection-system}: Bitalino, Pupil Head-Set, Haptic Feedback, OptiTrack, and Robot. Each device has designated application to acquire signals. For example, Bitalino provides ECG, EDA, and Accelerometer data and hand over these signals to LSL stream. Pupil Head-Set is a device that collects pupil dilation and gaze position, then this data can be sends to LSL stream. Other devices will follow similar approach to send data to the LSL.

While the subject is working with the robot, it is crucial to know the robot's speed, and the task started signal and task end signal for post signal processing. Labeling signal according to these events is vital for Machine Learning algorithm. 

\subsubsection{\textbf{Case 4: Understanding Human Comfort Level via Physiological Signal }}
\begin{figure}[ht]
	\centering
	\includegraphics[width=6.5cm,keepaspectratio]{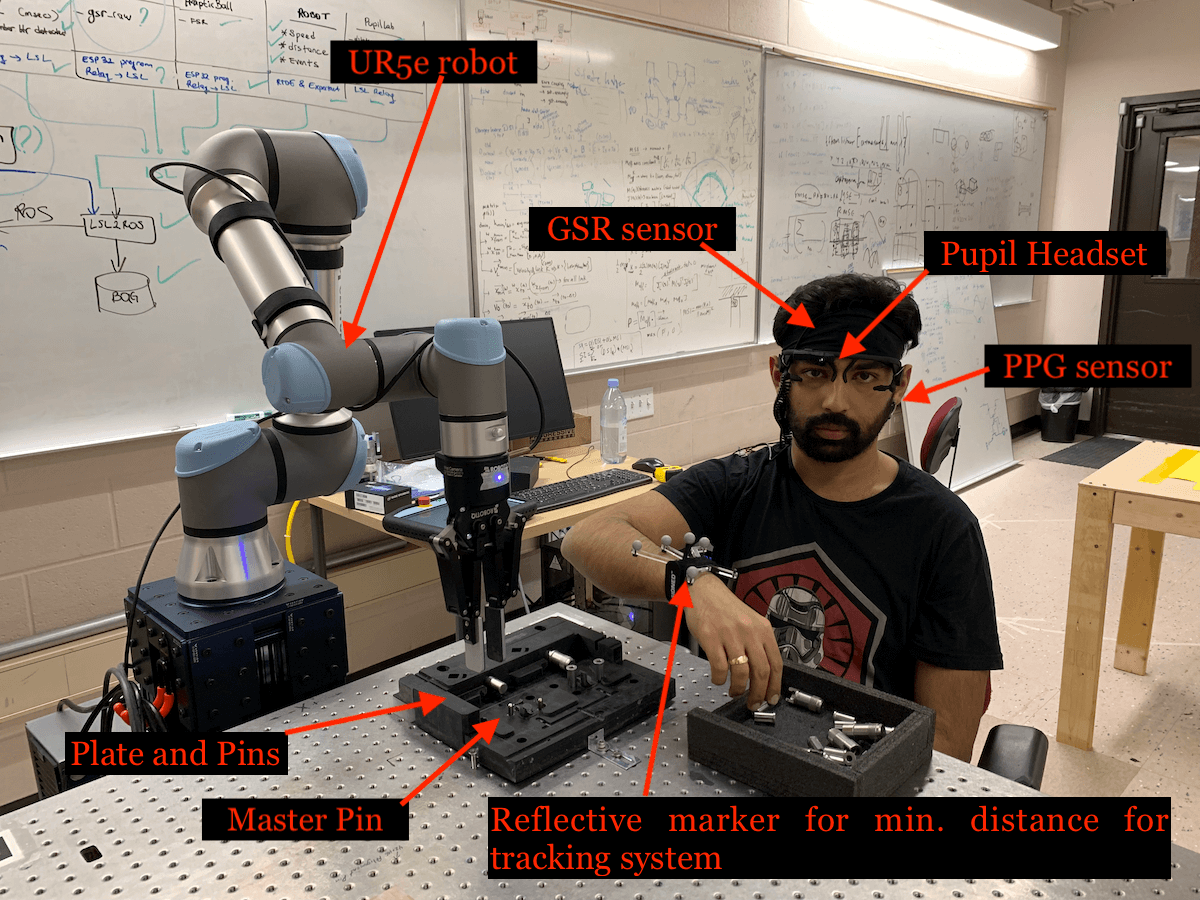}
	\caption{An experiment setup for monitoring human physiological signals during an HRC task to study human-comfort levels when the robot is moving at varying trajectory and accelerations.}
	\label{fig:case_study_4}
\end{figure}

The objective of the experiment is to monitor effect of the acceleration and trajectory of the robot on human physiological signals during collaborative task. By concurrently monitoring robot and human state, we hope to quantify a human operator comfort level while working in a shared workspace that may be triggered by changes in robot motion. The experiment was performed using UR5e (Universal Robot) six degree of freedom (DoF) arm robot, as shown in Figure \ref{fig:case_study_4}. The UR5e is a common collaborative robot with payload of 5 kg, which is suitable for manufacturing environment and laboratories. 

\section{Conclusion and  Future work}
\label{sec:FutureWork}
In this paper, an HRC-SoS framework is presented and three aspects of a Human Robot Collaboration setup :  \textit{Awareness}, \textit{Intelligence } and \textit{Compliance} are presented and discussed. Case studies and research highlighting the features  and these aspects has been presented. We believe a platform such as this will help in analyzing and building HRC setups that ensure human safety, build human trust in automation and optimize productivity.
 
\section*{Acknowledgment}
The authors are grateful to the staff of Multi Agent Bio-Robotics Laboratory (MABL) and the CM Collaborative Robotics Research (CMCR) Lab for their valuable inputs.

\bibliographystyle{IEEEtran}
\bibliography{CelalsBibtex,ShitijsBibtex1,ShitijsBibtex2,TulysBibtex,papers_v1.bib}

\end{document}